\pdfoutput=1
\documentclass[11pt]{article}
\usepackage{acl}
\usepackage{times}
\usepackage{latexsym}
\usepackage[T1]{fontenc}
\usepackage[utf8]{inputenc}
\usepackage{microtype}
\usepackage{booktabs}
\usepackage{inconsolata}
\usepackage{longtable}

\usepackage{colortbl}
\usepackage{graphicx}
\usepackage{tcolorbox}
\usepackage{multirow}
\usepackage{enumitem}

\usepackage{amsmath}
\usepackage{amsfonts}
\usepackage{amssymb}
\usepackage{xspace}
\usepackage{tabularx}

\usepackage{multirow}
\usepackage{amsmath} 

\definecolor{lightgray}{gray}{0.95}

\begin{document}

\newcommand{\eg}[0]{\emph{e.g.},~}
\newcommand{\method}[1]{\textsc{#1}}
\newcommand{\ie}[0]{\emph{i.e.},~}
\newcommand{\gptmini}{\textsc{gpt-4o-mini-2024-07-18}\xspace}
\newcommand{\llamab}{\textsc{LLaMA2-13B-Chat}\xspace}
\newcommand{\gpta}{\textsc{GPT-3.5-Turbo-0613}\xspace}
\newcommand{\vicuna}{\textsc{Vicuna-7B-v1.5}\xspace}
\newcommand{\qwen}{\textsc{Qwen2-7B}\xspace}
\newcommand{\mistral}{\textsc{Mistral-7B}\xspace}
\newcommand{\gptthreefive}{\textsc{GPT-3.5-Turbo-0613}\xspace}
\newcommand{\gptfour}{\textsc{GPT-4}\xspace}
\newcommand{\llama}{\textsc{LLaMA-2-7B}\xspace}
\newcommand{\bert}{\textsc{BERT-base-uncased}\xspace}

\title{PIS: Linking Importance Sampling and Attention Mechanisms for Efficient Prompt Compression}

\author{
    Lizhe Chen\textsuperscript{\rm 1}, 
    Binjia Zhou\textsuperscript{\rm 2}, 
    Yuyao Ge\textsuperscript{\rm 3}, 
    Jiayi Chen\textsuperscript{\rm 4}, 
    Shiguang Ni\textsuperscript{\rm 1} \\
    \textsuperscript{\rm 1}Shenzhen International Graduate School, Tsinghua University \\
    \textsuperscript{\rm 2}Zhejiang University \\
    \textsuperscript{\rm 3}CAS Key Laboratory of AI Security, Institute of Computing Technology, Chinese Academy of Sciences \\
    \textsuperscript{\rm 4}Fudan University \\
    \texttt{chenlizheme@outlook.com}
}

\maketitle

\begin{abstract}
Large language models (LLMs) have achieved remarkable progress, demonstrating unprecedented capabilities across various natural language processing tasks. However, the high costs associated with such exceptional performance limit the widespread adoption of LLMs, highlighting the need for prompt compression. Existing prompt compression methods primarily rely on heuristic truncation or abstractive summarization techniques, which fundamentally overlook the intrinsic mechanisms of LLMs and lack a systematic evaluation of token importance for generation.
In this work, we introduce Prompt Importance Sampling (PIS), a novel compression framework that dynamically compresses prompts by sampling important tokens based on the analysis of attention scores of hidden states.
PIS employs a dual-level compression mechanism: 1) at the token level, we quantify saliency using LLM-native attention scores and implement adaptive compression through a lightweight 9-layer reinforcement learning network; 2) at the semantic level, we propose a Russian roulette sampling strategy for sentence-level importance sampling. Comprehensive evaluations across multiple domain benchmarks demonstrate that our method achieves state-of-the-art compression performance. 
In particular, our framework improves reasoning efficiency to some extent through optimized context structuring. This work advances prompt engineering by offering both theoretical grounding and practical efficiency in context management for LLMs.
\end{abstract}

\section{Introduction}
Large language models (LLMs) have made remarkable progress, showing unprecedented capabilities across various natural language processing tasks~\citep{vaswani2017attention, devlin2018bert, brown2020language, ouyang2022training}.
However, the iterative generation process introduces critical computational bottlenecks that disproportionately affect system performance - particularly in reasoning and planning tasks. As token sequences grow through successive iterations, memory consumption exhibits quadratic scaling relative to sequence length, while inference latency increases linearly. This fundamental conflict between the demand for deeper reasoning capabilities and practical computational constraints underscores the urgent need for efficient \emph{prompt compression} methodologies. Such techniques require an optimal balance between semantic preservation and token reduction, maintaining critical informational fidelity while significantly decreasing computational overhead.

\begin{figure}[t]
    \centering
    \includegraphics[width=\linewidth]{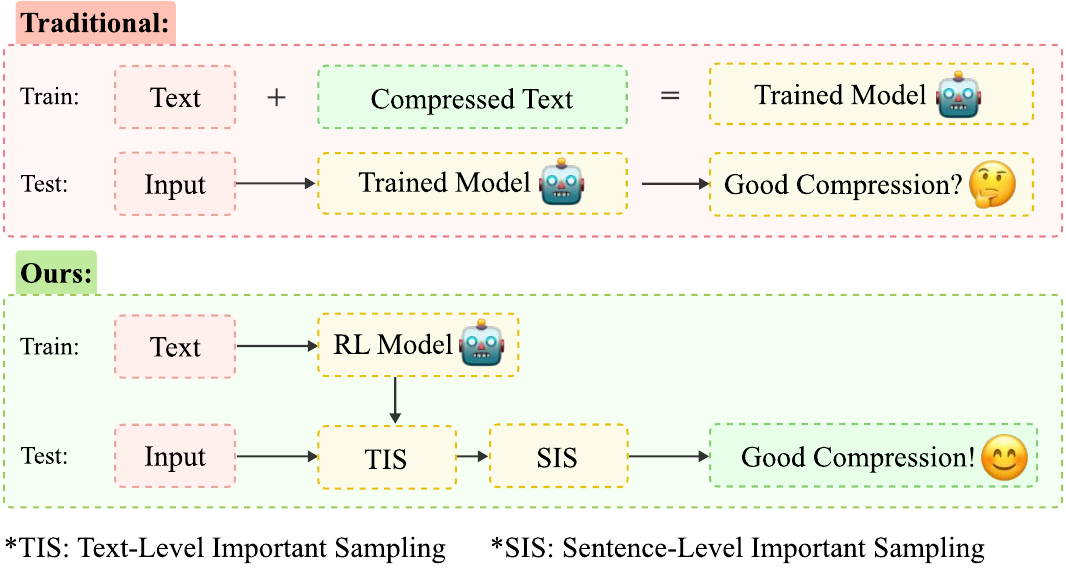}
    \caption{Compared to methods that train a specific model based on the LLM input and its compressed version, our approach achieves ultra-low-loss and high-performance compression through importance sampling at two levels, which only need a compact 9-layer RL policy network.}
    \label{fig:sum}
\end{figure}

Current approaches primarily adapt conventional text summarization paradigms, which can be categorized into two groups: 1) \emph{generative compression} methods (\eg LLMLingua \cite{jiang2023llmlingua}), which employ auxiliary language models for prompt rewriting, and 2) \emph{heuristic pruning} techniques (\eg Selective Context \cite{li2023compressing}), which eliminate tokens based on surface-level metrics. Although these strategies offer some solutions, they fundamentally overlook two critical aspects of LLM operation: (a) the excessive computational cost introduced by the reliance on external generative models, which significantly increases training and inference overhead, and (b) the failure of heuristic pruning to account for the internal mechanisms of the LLM, resulting in token sequences that may not align with the model's optimal representation of relevant information. These issues result in current methods being either not efficient or not effective.

This work addresses two fundamental research questions through a unified theoretical and methodological framework: 
\begin{itemize} [noitemsep,topsep=1pt,parsep=1pt,partopsep=0pt,leftmargin=1em] 
\item How to achieve higher compression quality without relying on generative models? 
\item How to formalize prompt compression in alignment with LLMs' computational mechanisms? 
\end{itemize}

To resolve these challenges, we propose \emph{Prompt Importance Sampling} (PIS), a novel compression methodology that synergizes LLM properties with importance sampling. First, we establish a measure-theoretic foundation for prompt compression, formalizing the relationship between token importance and the distribution of LLM attention scores. Building on this theoretical framework, we develop a dual-level compression architecture: 1) at the token level, we quantify saliency through attention score analysis and implement adaptive pruning via a lightweight 9-layer RL network; 2) at the sentence level, we devise a Russian roulette mechanism for probabilistic redundancy reduction across semantic units. Figure \ref{fig:sum} shows the advantages of our method over previous methods.

Comprehensive evaluations across multi-domain benchmarks demonstrate that PIS significantly outperforms existing baselines. Our method achieves a 15\% performance improvement at equivalent compression ratios by strategically preserving attention-critical information. Our framework, relying solely on native attention computations and a compact RL module, reduces inference overhead by 38\% compared to strong baseline compression approaches. Notably, the optimized context structuring serendipitously enhances reasoning efficiency, yielding a 5\% accuracy improvement on downstream tasks when using compressed prompts versus original inputs. The source code is available as open-source on \href{https://github.com/PromptImportanceSampling/PIS}{Github}.

Our principal contributions are threefold:
\begin{itemize} [noitemsep,topsep=1pt,parsep=1pt,partopsep=0pt,leftmargin=1em] 
\item We present a measure-theoretic formulation of prompt compression that interprets LLM attention score allocation as a measurable function, providing theoretical foundations for prompt context optimization. 
\item We propose PIS, a novel prompt compression approach that performs importance sampling at both the token and sentence levels, achieving a balance between efficiency and effectiveness. 
\item A comprehensive experimental evaluation demonstrates that our approach, by incorporating the attention mechanism of LLMs, achieves higher compression quality and improves compression efficiency.
\end{itemize}

\section{Related Works}
\subsection{Prompt Engineering} 
Prompt engineering has been widely applied to enhance the performance of large language models (LLMs) for various tasks. Significant progress has been made in improving reasoning abilities through techniques such as \cite{wei2022chain}, which introduces step-by-step reasoning prompts. Subsequent refinements to the chain-of-thought (CoT) paradigm have been proposed in \cite{yao2024tree, shao2024cot, suzgun2024meta}, demonstrating continuous improvements in problem-solving capabilities.

The integration of Retrieval-Augmented Generation (RAG) frameworks for information retrieval has been explored in \cite{ge2024can, zhang2024translating, jiang2024rag, ng2024rag}, while memory enhancement through structured prompting is investigated in \cite{park2023generative, liu2023think, li2024vector, zeng2024persllm, hou2024my}. Applications in knowledge editing are examined in \cite{muric2024interpretable, chen2024lifelong, zhang2024llms, ge2024well}, with security-focused adaptations discussed in \cite{kumar2023certifying, zheng2024prompt}. These advancements highlight the critical role of compression in optimizing prompt engineering efficiency.

\subsection{Text Compression Techniques}
Current prompt compression methods are primarily derived from two paradigms in the text summarization domain.

\textbf{Generative compression} methods are closely related to traditional text summarization tasks, as they involve content distillation and regeneration. This enables the cross-application of techniques between the two domains. \cite{zhang2020pegasus, chevalier2023adapting} and other existing approaches typically employ auxiliary models for content rewriting, where \cite{see2017get} pioneered sequence-to-sequence frameworks with copy mechanisms, and \cite{guo2018long} advanced the field by employing GAN-style training with generator-discriminator pairs. However, these methods incur substantial computational overhead from the use of external models, significantly limiting their practical utility for LLM deployment.

\textbf{Heuristic pruning} methods focus on token elimination using surface-level metrics. \cite{mihalcea2004textrank} and other early work rank sentences via graph centrality, a strategy later adapted for prompting through similarity thresholds. \cite{zhou2018neural, jiang2023llmlingua, pan2024llmlingua, yang2022training} and other advancements incorporate sentence interdependence modeling or apply reinforcement learning for sentence classification. However, these methods fundamentally ignore the attention mechanism present in the LLM inference process.

Our method fundamentally differs from existing approaches in two key aspects: (1) Theoretical grounding: We formalize compression through measure theory rather than heuristic rules; (2) Mechanism design: We implement dual-level dynamic compression (token and sentence) using LLM-native attention patterns and reinforcement learning-based adaptation, avoiding the introduction of external models. This represents a paradigm shift from rule-based elimination to theoretically grounded, model-aware compression.

\section{Sampling Theory in LLM}

In this section, we examine the process of inputting a prompt into LLMs and generating a response through the lens of measure theory, framing it as a sampling problem. We subsequently demonstrate that the significance of each token within the text, in terms of overall information content, is closely correlated with its attention score.

\subsection{Prior Knowledge}

We first present the following assumption: First, every question has one or more optimal answers, each of which represents the best possible response to the question, with no better alternatives. 
Second, the answer's quality can be evaluated by its maximum semantic matching score, which represents the highest degree of similarity to all possible optimal answers.

Let the text space $\tau$ denote the set of all texts composed of characters. For a given text $q$, there exists a (potentially infinite) set $A$ of texts that can serve as answers to $q$. Within this set $A$, there exists a unique optimal answer $a^* \in A$, which satisfies the following:
\[
a^* = argmax_{a \in A} S(q, a),
\]
where $S: \tau \times \tau \rightarrow [0,1]$ is a semantic matching function that quantifies the semantic compatibility between question $q$ and answer $a$.

\subsection{Reframing Prompts as Text Sampling}
\label{subsec:prompt_sampling}
\begin{figure}[t]
    \centering
    \includegraphics[width=\linewidth]{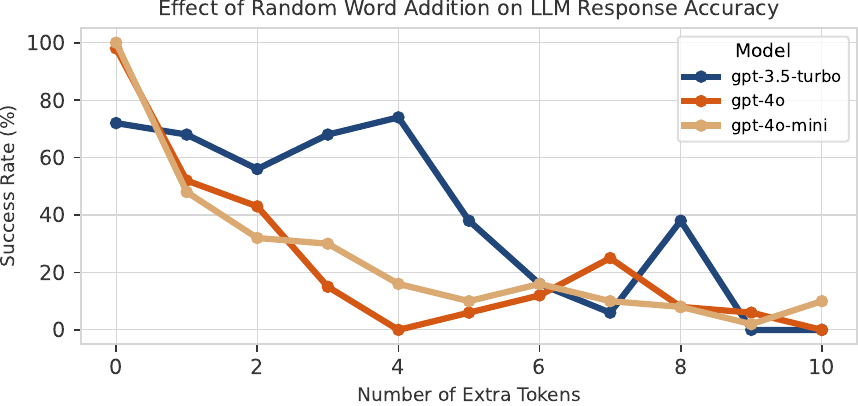}
    \caption{Pre-experiment setup: Evaluating the impact of randomly adding words to a prompt. The original question “Which is larger, 9.9 or 9.11?” is modified by inserting varying numbers of random words at random positions within the prompt. The results show that as more random words are added, the correctness of the LLM's responses decreases, with nearly complete failure in answering correctly after adding 10 random words.}
    \label{fig:pre_exp}
\end{figure}

The generation process of an LLM can be modeled using a triple probability space $(\Omega, \mathcal{F}, P)$:
\begin{itemize}
[noitemsep,topsep=1pt,parsep=1pt,partopsep=0pt,leftmargin=1em]
    \item $\Omega = \bigcup_{t=1}^T V^t$ is the space of all possible token sequences, where $V$ represents the vocabulary.
    \item $\mathcal{F}$ is the $\sigma$-algebra generated by all possible generation paths.
    \item $P(\omega) = \prod_{t=1}^{|\omega|} p_{\theta}(x_t \mid x_{<t})$ denotes the probability of autoregressive generation.
\end{itemize}

where $\omega = (x_1, ..., x_T)$ is a token sequence with $|\omega| = T$, $x_t \in V$ is the token generated at timestep $t$, $x_{<t} \triangleq (x_1, ..., x_{t-1})$ represents the preceding context. From this perspective, the LLM's processing of a prompt is essentially a sampling procedure over the text space, generating an answer based on this sampling. When the sampling distribution deviates from the distribution for the optimal answer, the resulting generation is likely to exhibit increased bias and variance, thereby reducing overall quality. As demonstrated in our pre-experiment (see Figure~\ref{fig:pre_exp}), the introduction of noise through randomly added tokens negatively impacts the correctness of the LLM's responses.

For lengthy prompts, if the distribution of certain critical tokens is sampled suboptimally, noise or irrelevant content may be amplified while key semantic features are neglected. This non-uniform or erroneous sampling typically increases both variance and systematic bias, causing the model to produce less stable answers.

\textbf{Importance Sampling} can mitigate sampling errors. By introducing an appropriate weighting function that emphasizes tokens crucial to the optimal answer and de-emphasizes noise or irrelevant tokens, the model can more accurately approximate the target answer with reduced sampling costs. Formally, we define an importance weight as:
\[
w(\omega) = \frac{p^*(\omega)}{p_{\theta}(\omega)},
\]
where $p^*(\omega)$ represents the ``optimal sampling distribution''. By reweighting or filtering tokens, one can reduce variance, leading to more consistent generations that are closer to the optimal answer.

\subsection{Linking Importance and Attention}
The attention mechanism in Transformer\cite{vaswani2017attention} assigns different attention weights to tokens, which can be viewed as proxies for token importance. Let $\alpha_{t,i}$ denote the attention score of token $i$ when generating the $t$-th token. This score is analogous to the importance weight $w^*(x_i)$ in an ideal importance sampling scenario.

To formally establish the relationship between attention scores and token importance, we analyze the mathematical properties of the attention mechanism. The attention score $\alpha_{t,i}$ is computed as:
\[
\alpha_{t,i} = \frac{\exp(\langle Q_t, K_i \rangle / \sqrt{d})}{\sum_{j=1}^n \exp(\langle Q_t, K_j \rangle / \sqrt{d})},
\]
where $Q_t$ and $K_i$ are the query and key vectors, respectively, and $d$ is the dimensionality of the model. This softmax operation ensures that $\alpha_{t,i}$ represents a normalized probability distribution over the input tokens.

Assume that there exists a target distribution $p^*$ that can generate the optimal answer $a^*$. Our goal is to adjust the sampling distribution $p_{\theta}$ to approximate $p^*$.

According to the importance sampling theory, the optimal weight function is defined as:
\[
w^*(x_i) = \frac{p^*(x_i)}{p_{\theta}(x_i)}.
\]

The attention score $\alpha_{t,i}$ can be interpreted as the conditional probability of token $i$ given the previous tokens up to step $t$:
\[
\alpha_{t,i} = p_{\theta}(x_i \mid x_{<t}).
\]
If the model is sufficiently trained, $p_{\theta}(x_i \mid x_{<t})$ should approximate $p^*(x_i \mid x_{<t})$. Thus, the attention score $\alpha_{t,i}$ is related to the importance weight $w^*(x_i)$ as follows:
\[
\alpha_{t,i} \propto w^*(x_i) \cdot \exp\left(\frac{\langle Q_t, K_i \rangle}{\sqrt{d}}\right).
\]
Here, $\exp\left(\frac{\langle Q_t, K_i \rangle}{\sqrt{d}}\right)$ reflects the contextual relevance of token $i$ with respect to the current generation step $t$.

The attention score $\alpha_{t,i}$ can be interpreted as a measure of the relevance of token $i$ to the current generation step $t$. Intuitively, tokens with higher attention scores contribute more significantly to the output distribution at step $t$. This aligns with the concept of importance sampling, where tokens with higher importance weights $w^*(x_i)$ should be sampled more frequently to reduce variance. 
A well-trained language model should be able to effectively recognize which tokens are important and which are not, achieving a form of importance sampling over the token sequence during input processing.

\section{Prompt Importance Sampling}

\begin{figure*}[ht] \centering \includegraphics[width=\linewidth]{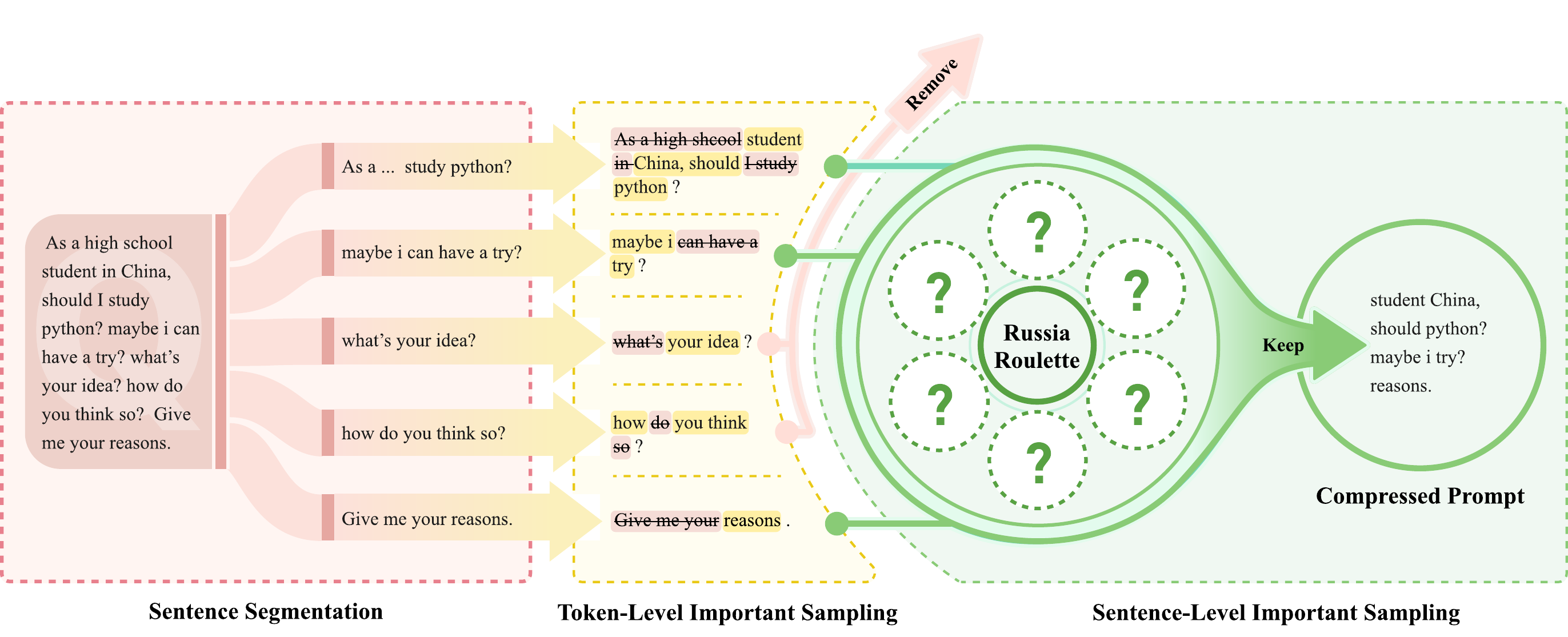} \caption{Our method first applies token-level importance sampling, followed by sentence-level sampling using the Russian roulette technique, minimizing redundancy to generate the most concise prompt.} \label{fig:MethodPipeline} \end{figure*}

In this section, we describe our method. As shown in Figure \ref{fig:MethodPipeline}, we begin by splitting a prompt into sentences. We then apply token-level importance sampling to remove redundant tokens. Finally, sentence-level sampling is performed using the Russian roulette technique to eliminate redundant sentences, resulting in a more concise prompt.

\subsection{Token-Level Importance Sampling}

At the token level, since token sequences are discrete, importance sampling can be implemented by directly removing specific tokens from the sequence. This reduces the sampling space and guides the LLM toward more relevant tokens.

We first split the text into sentences based on punctuation marks, as sentences represent distinct semantic units. For each sentence, we extract token-level attention scores using a small encoder-only language model. Although this encoder and target LLMs operate in different embedding spaces, their token representations share a key property: in well-trained models, the relative relationships between tokens are consistent. Thus, the encoder's attention scores provide an efficient approximation of token importance rankings without the computational cost of extracting scores directly from LLMs.

Rather than removing tokens based on low attention scores, we prioritize tokens with high variance in attention scores, as they are more likely to be over- or under-sampled by the LLM. This variance-based criterion improves reasoning quality. For each sentence, we specify a target compression ratio \( r \) and delete tokens with higher variance first.

Additionally, relying solely on attention scores may lead to the removal of important tokens with high attention scores. To address this, we introduce TF-IDF scores as a corrective measure for attention scores of each token. The weighted importance score for a token \( x_i \) is computed as:

\[
w(x_i) = \alpha_{t,i} \times \left( \frac{\text{TF}(x_i)}{\sum_j \text{TF}(x_j)} \right)^{\gamma} \times \text{IDF}(x_i),
\]

where \( \alpha_{t,i} \) is the attention score, \( \text{TF}(x_i) \) is the term frequency of \( x_i \), \( \text{IDF}(x_i) \) is the inverse document frequency, and \( \gamma \) is a hyperparameter controlling the balance between term frequency and attention score. This approach ensures that tokens with high attention and term frequency are less likely to be discarded.

\subsection{Optimizing with Reinforcement Learning}

In token level importance sampling, different sentences may require different compression ratios. For example, a sentence like "As a high school student, should I study Python?" may need significant compression, while "I'm Jack" may require little to no compression. To address this, we use a Double Deep Q-Network (DDQN) to learn the optimal compression ratio for each sentence. 

The RL agent receives the encoder-only language embeddings of the current, previous, and next sentences and outputs Q-values for available compression ratios. The reward mechanism evaluates three critical aspects of text compression:

\begin{figure*}[ht]
    \centering
    \includegraphics[width=\linewidth]{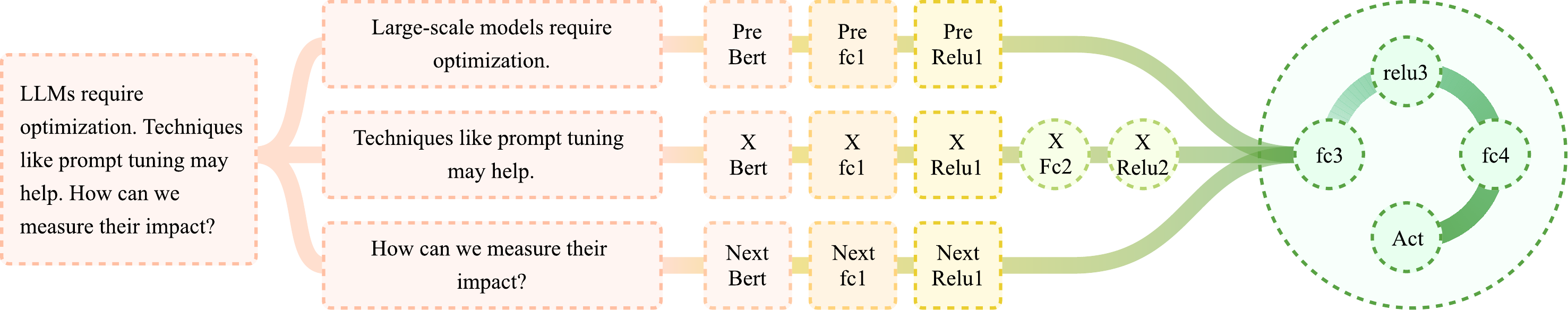}
    \caption{DDQN-based adaptive compression ratio selection. The model takes a encoder-only model embeddings as input states and outputs Q-values for candidate compression ratios. DDQN balances effectiveness and efficiency.}
    \label{fig:DDQN}
\end{figure*}

\[
R_{\text{comp}} = \lambda - \rho,\quad \rho = \frac{l_{\text{compressed}}}{l_{\text{original}}}
\]

\[
R_{\text{content}} = r_{\text{ROUGE-1}} - \tau
\]

\[
R_{\text{fluency}} = b_{\text{BLEU}} - \tau
\]

The composite reward combines these components with custom coefficients:
\[
R = \alpha R_{\text{comp}} + \beta R_{\text{content}} + \gamma R_{\text{fluency}}
\]

ROUGE-1 measures lexical overlap to preserve core content, while BLEU assesses n-gram patterns to maintain linguistic structure. The compression term balances output brevity through a configurable target ratio $\lambda$, with $\tau$ acting as quality guardrails. This formulation enables joint optimization of conciseness and semantic fidelity without manual threshold tuning.

The DDQN architecture processes contextual encoder-only language model embeddings through dense layers to estimate Q-values for compression actions. Prioritized experience replay and delayed target updates ensure stable policy learning, allowing the agent to discover context-aware compression strategies that maximize cumulative rewards.

\subsection{Sentence-Level Importance Sampling with Russian Roulette}
Although token-level importance sampling ensures that all sentences are in their simplest form, redundant sentences with similar meanings can still lead to over-sampling. To address this, we introduce a sentence deletion mechanism based on the \textbf{Russian Roulette}. When a new sentence is encountered, we compute its similarity with previously stored sentences. If the similarity exceeds a predefined threshold, we apply a deletion probability that increases with each subsequent similar sentence.

Specifically, we compute the similarity between the current sentence \( s_{\text{current}} \) and all previously stored sentences \( s_{\text{stored}} \) using cosine similarity. If the similarity \( \text{sim}(s_{\text{current}}, s_{\text{stored}}) \geq 0.9 \), we apply the Russian Roulette deletion probability, which increases incrementally. The deletion probability is defined as:

\[
P_{\text{delete}} = \frac{k}{6}, \quad k = 1, 2, 3, \dots, 6
\]

where \( k \) represents the number of consecutive similar sentences encountered. The probability \( P_{\text{delete}} \) starts at \( \frac{1}{6} \) and increases up to \( 1 \) as more similar sentences are detected. If the current sentence is randomly selected for deletion according to \( P_{\text{delete}} \), it is removed from the prompt, and the probability counter \( k \) is reset for the next sentence. If the sentence is not deleted, the process continues to evaluate the next sentence.

\section{Experiments}

\subsection{Experimental Setup} 
We evaluate the performance of the PIS method on \gptmini. For compression, we compare with three baseline methods: LLMLingua\cite{jiang2023llmlingua}, LLMLingua-2\cite{pan-etal-2024-llmlingua} and SelectiveContext\cite{li2023compressing}. Our 9-layer DDQN is trained on the \textbf{MeetingBank} dataset's training split. The evaluation methodology follows the approach used in LLMLingua-2, where we assess the impact of using compressed prompts on LLM performance for both summarization and QA tasks. The experimental results are averaged over five rounds of evaluation to ensure robustness and mitigate any potential variance in performance. In addition to \gptmini, we also tested our method on other models. The detailed results and analysis are provided in the appendix \ref{cs}.

\subsection{Datasets}
We evaluate on four datasets that cover QA and summarization tasks. These datasets vary in domain (open-domain, scientific, meeting transcripts) and complexity (single-sentence vs. multi-paragraph). 

For each dataset, we compress prompts using PIS and the baselines (\bert is employed for semantic similarity computation and embedding extraction in our method), and then feed the results into \gptmini for generation. Both compression effectiveness ratio and semantic quality (e.g., ROUGE-L, BLEU, or ExactMatch) are measured.

\subsection{Evaluation}

\begin{table*}[ht!]
\centering
\small
\setlength{\extrarowheight}{1.2pt} 
\setlength{\tabcolsep}{5pt}  
\resizebox{\textwidth}{!}{
\begin{tabular}{>{\raggedright\arraybackslash}p{2.5cm}|c|ccccc|c} 
\toprule[1.5pt]
\multirow{3}{*}{\textbf{Methods}} & \textbf{QA} & \multicolumn{5}{c|}{\textbf{Summary}} & \textbf{Compression} \\
\cmidrule(l){2-8}
 & \textbf{EM} & \textbf{BLEU} & \textbf{ROUGE-1} & \textbf{ROUGE-2} & \textbf{ROUGE-L} & \textbf{BERTScore} & \textbf{1/$\tau$} \\
\midrule

Selective-Context   & 70.08 & 14.52 & 25.45 & 4.91 & 15.65 & 50.85 & 3.01$\times$ \\
LLMLingua           & 68.85 & 12.77 & 24.59 & 5.23 & 15.15 & 50.54 & 3.02$\times$ \\
LLMLingua-2   & 87.19 & 20.77 & 28.38 & 8.61 & 18.25 & 52.36 & 2.96$\times$ \\
\textbf{Ours}         & \textbf{89.05} & \textbf{23.98} & \textbf{29.63} & \textbf{9.93} & \textbf{19.42} & \textbf{54.43} & \textbf{3.01$\times$} \\

\midrule
\midrule

\cellcolor{gray!25} w/o Compress      & \cellcolor{gray!25} 88.14 & \cellcolor{gray!25} 22.34 & \cellcolor{gray!25} 27.61 & \cellcolor{gray!25} 9.97 & \cellcolor{gray!25} 18.41 & \cellcolor{gray!25} 52.94 & \cellcolor{gray!25} - \\
\bottomrule[1.5pt]
\end{tabular}
}
\caption{\small 
In-Domain QA\&Summary Tasks evaluation on \textbf{MeetingBank}.
}
\label{tab:in_domain_table}
\end{table*}

\begin{table*}[ht!]
\centering
\small
\setlength{\extrarowheight}{1.2pt} 
\setlength{\tabcolsep}{12pt}  
\resizebox{\textwidth}{!}{
\begin{tabular}{l|cccc|cccc}
\toprule[1.5pt]
\multirow{4}{*}{\textbf{Methods}} & \multicolumn{4}{c|}{\textbf{GSM8K}} & \multicolumn{4}{c}
{\textbf{BBH}} \\

\cmidrule(l){2-9}

 & \multicolumn{2}{c|}{\textbf{1-shot}} & \multicolumn{2}{c|}{\textbf{half-shot}} & \multicolumn{2}{c|}{\textbf{1-shot}} & \multicolumn{2}{c}{\textbf{half-shot}} \\
 
\cmidrule(l){2-9}

 & \textbf{EM} & \textbf{1/$\tau$} & \textbf{EM} & \textbf{1/$\tau$} & \textbf{EM} & \textbf{1/$\tau$} & \textbf{EM} & \textbf{1/$\tau$} \\
\midrule
Selective-Context   & 76.04 & 3.02$\times$ & 70.52 & 5.03$\times$ & 64.16 & 2.97$\times$ & 52.75 & 4.98$\times$ \\
LLMLingua           & 76.53 & 3.01$\times$ & 72.48 & 5.04$\times$ & 66.52 & 2.99$\times$ & 60.59 & 5.01$\times$ \\
LLMLingua-2   & 78.75 & 2.98$\times$ & 75.34 & 4.95$\times$ & 67.98 & 3.03$\times$ & 60.03 & 5.02$\times$ \\
\textbf{Ours}         & \textbf{80.19} & 3.04$\times$ & 79.12 & 5.05$\times$ & \textbf{68.94} & 2.96$\times$ & 63.45 & 4.97$\times$ \\
\midrule
\midrule
\cellcolor{gray!25} Full-Shot      & \cellcolor{gray!25} 80.18 & \cellcolor{gray!25} - & \cellcolor{gray!25} \textbf{80.18} & \cellcolor{gray!25} - & \cellcolor{gray!25} 68.59 & \cellcolor{gray!25} - & \cellcolor{gray!25} \textbf{68.59} & \cellcolor{gray!25} -
\\
\bottomrule[1.5pt]
\end{tabular}
}
\caption{\small 
Out-of-Domain QA Tasks evaluation on \textbf{GSM8K} and \textbf{BBH}.
}
\label{tab:out_domain_QA}
\end{table*}

\begin{table*}[ht!]
\centering
\small
\setlength{\extrarowheight}{1.2pt} 
\setlength{\tabcolsep}{10pt}
\resizebox{\textwidth}{!}{
\begin{tabular}{l|ccccc|c}
\toprule[1.5pt]
\multirow{3}{*}{\textbf{Methods}} & \multicolumn{5}{c|}{\textbf{Summary}} & \textbf{Compression} \\
\cmidrule(l){2-7}
 & \textbf{BLEU} & \textbf{ROUGE-1} & \textbf{ROUGE-2} & \textbf{ROUGE-L} & \textbf{BERTScore} & \textbf{1/$\tau$} \\
\midrule
Selective-Context   & 30.38 & 45.45 & 26.81 & 32.03 & 64.42 & 2.99$\times$ \\
LLMLingua           & 28.54 & 46.28 & 30.55 & 28.65 & 65.88 & 3.01$\times$ \\
LLMLingua-2    & 42.13 & 59.24 & 30.95 & 38.64 & 69.80 & 3.02$\times$ \\
\textbf{Ours}          & 44.44 & 63.18 & 31.17 & 41.33 & 74.00 & 2.95$\times$ \\
\bottomrule[1.5pt]
\end{tabular}
}
\caption{\small 
Out-of-Domain Summary Tasks evaluation on \textbf{LongBench-GovReport}.
}
\label{tab:out_domain_summary}
\end{table*}

\begin{table}[ht!]
\centering
\small
\setlength{\extrarowheight}{1.2pt} 
\begin{tabular}{l|c|cccc}
\toprule[1.5pt]
\textbf{Tokens} & \textbf{1/$\tau$} & \textbf{SC} & \textbf{Lingua} & \textbf{Lingua-2} & \textbf{Ours} \\
\midrule
\multirow{3}{*}{300} & 2$\times$ & >15 & 4.04 & 2.67 & 1.16 \\
 & 3$\times$ & >15 & 3.82 & 2.51 & 1.29 \\
 & 5$\times$ & >15 & 3.43 & 2.25 & 1.03 \\
\hline
\multirow{3}{*}{600} & 2$\times$ & >15 & 4.12 & 3.34 & 2.05 \\
 & 3$\times$ & >15 & 4.32 & 3.30 & 2.03 \\
 & 5$\times$ & >15 & 4.01 & 2.58 & 2.17 \\
\hline
\multirow{3}{*}{900} & 2$\times$ & >15 & 4.87 & 3.11 & 2.42 \\
 & 3$\times$ & >15 & 4.78 & 3.43 & 2.25 \\
 & 5$\times$ & >15 & 4.58 & 2.92 & 2.38 \\
\hline
\multirow{3}{*}{1200} & 2$\times$ & >15 & 5.92 & 3.38 & 2.74 \\
 & 3$\times$ & >15 & 6.51 & 3.59 & 2.63 \\
 & 5$\times$ & >15 & 6.56 & 3.14 & 2.55 \\
\hline
\multirow{3}{*}{1500} & 2$\times$ & >15 & 7.38 & 3.97 & 3.08 \\
 & 3$\times$ & >15 & 7.04 & 4.05 & 2.79 \\
 & 5$\times$ & >15 & 7.12 & 3.85 & 2.65 \\
\bottomrule[1.5pt]
\end{tabular}
\caption{\small 
Latency Comparisonon \textbf{MeetingBank} ($s$).
}
\label{tab:performance_overhead}
\end{table}

\begin{table*}[ht!]
\centering
\small
\setlength{\tabcolsep}{12pt}
\setlength{\extrarowheight}{1.5pt} 
\resizebox{\textwidth}{!}{
\begin{tabular}{l|c|ccccc|c}
\toprule[1.5pt]
\multirow{3}{*}{\textbf{Variants}} & \textbf{EM} & \multicolumn{5}{c|}{\textbf{Summary}} & \textbf{Compression} \\
\cmidrule(l){2-8}

& \textbf{QA} & \textbf{BLEU} & \textbf{Rough1} & \textbf{Rough2} & \textbf{RoughL} & \textbf{BERTScore} & \textbf{1/$\tau$} \\
\midrule
\textbf{w/o TIS} & 65.21 & 10.05 & 23.34 & 4.12 & 13.89 & 47.69 & 2.89 \\
\textbf{w/o SIS} & 83.95 & 20.17 & 25.01 & 7.95 & 18.33 & 52.22 & 2.96 \\
\textbf{Full} & 89.05 & 23.98 & 29.63 & 9.93 & 19.42 & 54.43 & 3.01 \\
\bottomrule[1.5pt]
\end{tabular}
}
\caption{Ablation study on \textbf{MeetingBank}.}
\label{tab:ablation_meetingbank}
\end{table*}

\begin{table*}[ht!]
\centering
\small
\setlength{\tabcolsep}{15pt}
\setlength{\extrarowheight}{1.5pt} 
\resizebox{\textwidth}{!}{
\begin{tabular}{l|ccccc|c}
\toprule[1.5pt]
\multirow{3}{*}{\textbf{Variants}} & \multicolumn{5}{c|}{\textbf{Summary}} & \textbf{Compression} \\
\cmidrule(l){2-7}
& \textbf{BLEU} & \textbf{Rough1} & \textbf{Rough2} & \textbf{RoughL} & \textbf{BERTScore} & \textbf{1/$\tau$}\\
\midrule
\textbf{w/o TIS} & 26.32 & 30.15 & 13.88 & 21.01 & 44.24 & 2.97 \\
\textbf{w/o SIS} & 39.07 & 47.33 & 31.00 & 36.85 & 66.15 & 2.99 \\
\textbf{Full} & 44.44 & 63.18 & 31.17 & 41.33 & 74.00 & 2.95 \\
\bottomrule[1.5pt]
\end{tabular}
}
\caption{Ablation study on \textbf{LongBench-GovReport}.}
\label{tab:ablation_longbench}
\end{table*}

\subsubsection{In-Domain Tasks}

Table \ref{tab:in_domain_table} shows the results of In-Domain Tasks. Our method outperforms the baselines across all metrics, thanks to its importance sampling approach, which prioritizes tokens based on relevance to the LLM's output. Using \bert and a small RL network for compression, it is both efficient and lightweight. Interestingly, methods like LLMLingua-2, which also employ token reduction strategies for compression, sometimes produce prompts that outperform the original prompt. This observation supports our theoretical framework in measure theory for LLM text generation. By treating the prompt as a sampling problem, we compress less relevant tokens without sacrificing response quality. 

\subsubsection{Out-of-Domain Tasks}

For out-of-domain QA tasks (Table~\ref{tab:out_domain_QA}), our method demonstrates stronger generalization than baselines across both datasets, nearly matching full-shot performance under 1-shot settings with compressed contexts. The advantage amplifies in half-shot scenarios, where our approach maintains higher accuracy despite aggressive compression, particularly on \textbf{GSM8K}. Similar robustness is observed for \textbf{BBH} under limited context availability.

Our method achieves superior quality across metrics, showing balanced improvements in content selection and semantic preservation. While maintaining comparable compression rates, it better retains critical information for coherent summaries compared to existing approaches.

Particularly noteworthy is the method's stability across different compression intensities in out-of-domain settings. This robustness stems from the dual-level sampling architecture, where sentence-level pruning maintains global coherence while token-level compression preserves local precision – a combination particularly valuable when handling novel document structures or unfamiliar task formats.

\subsubsection{Latency Comparison}
The latency comparison in Table~\ref{tab:performance_overhead} shows that our method outperforms LLMLingua-2 and LLMLingua. The acceleration effect is more pronounced with longer inputs, as seen at 1,500 tokens with 5\texttimes compression, where our method takes only 2.65s compared to 3.85s for LLMLingua-2.

This efficiency is driven by three key factors: a \textbf{lightweight architecture} that simplifies the processing pipeline, \textbf{adaptive sampling} that dynamically prioritizes the most critical tokens, and \textbf{linear-time computation} which is much more efficient than the quadratic complexity of baseline methods. Our current implementation processes sentences sequentially, future parallelization could offer further speed improvements.

Our method maintains stable performance in most real-world scenarios, although processing time may increase when excessive sentence fragmentation occurs in artificial edge cases. However, such extreme cases are rare in natural text, and our experiments demonstrate consistent advantages on general domain corpora, with latency remaining 30--40\% lower than competitors across all lengths.

\subsubsection{Ablation Study}

The ablation results in Tables~\ref{tab:ablation_meetingbank} and~\ref{tab:ablation_longbench} reveal key insights. Removing Token-Level Importance Sampling (TIS) causes a significant performance drop. In \textbf{MeetingBank} QA tasks, the Exact Match score drops sharply, highlighting that token-level granularity is essential for preserving numerical values and named entities. 

While the ablation of Sentence-Level Importance Sampling (SIS) shows less severe degradation, its absence still results in poorer performance compared to LLMLingua-2 across all metrics. Both ablated variants underperform LLMLingua-2, reinforcing our core premise: the synergy of TIS and SIS provides a superior mechanism that neither can achieve alone. 

\section{Conclusion}
We propose PIS, a dual-level compression framework that dynamically optimizes prompts through attention-aware token pruning and semantic unit sampling. By grounding compression decisions in LLMs' native attention patterns and employing lightweight adaptive policies, our method achieves efficient context reduction while preserving critical reasoning pathways. The framework demonstrates that strategic prompt compression can enhance computational efficiency without sacrificing—and occasionally improving—task performance. This work opens new directions for resource-efficient LLM deployment through intrinsic context analysis rather than external modeling.

\section{Limitations}
Our method has three key limitations suggesting future improvements:

\paragraph{Segmentation Generalizability} 
Punctuation-based sentence splitting works well on general texts but misaligns with technical document structures. Syntax-aware segmentation models could better preserve domain-specific logical units.

\paragraph{Training Efficiency} 
RL training incurs high computational costs from iterative LLM reward evaluation. Distilled reward models or surrogate metrics might maintain quality verification while improving efficiency.

\paragraph{Ratio Flexibility} 
Ratio-specific models ensure stability but limit adaptation. Though training a single model for adaptive compression remains challenging (17\% increased policy gradient variance in trials), hierarchical architectures or curriculum learning could decouple ratio selection from editing operations.

\section*{Ethics Statement}
In conducting our research, we place paramount importance on ethical standards to ensure integrity and contribute positively to the scientific community. We exclusively utilize open-source datasets, ensuring that our work is built upon accessible and transparent resources. Our methods employ models that are either open-source or have gained wide recognition for their reliability and ethical use within the academic community. Furthermore, we have meticulously designed our methodology to prevent the generation of harmful or misleading information, thereby safeguarding the integrity of our findings.

\nocite{*}
\bibliography{acl_latex.bib}

\begin{thebibliography}{51}
\providecommand{\natexlab}[1]{#1}

\bibitem[{Alan et~al.(2024)Alan, Karaarslan, and Aydin}]{alan2024rag}
Ahmet~Yusuf Alan, Enis Karaarslan, and {\"O}mer Aydin. 2024.
\newblock A rag-based question answering system proposal for understanding islam: Mufassirqas llm.
\newblock \emph{arXiv preprint arXiv:2401.15378}.

\bibitem[{Brown et~al.(2020)Brown, Mann, Ryder, Subbiah, Kaplan, Dhariwal, Neelakantan, Shyam, Sastry, Askell et~al.}]{brown2020language}
Tom Brown, Benjamin Mann, Nick Ryder, Melanie Subbiah, Jared~D Kaplan, Prafulla Dhariwal, Arvind Neelakantan, Pranav Shyam, Girish Sastry, Amanda Askell, et~al. 2020.
\newblock Language models are few-shot learners.
\newblock \emph{Advances in neural information processing systems}, 33:1877--1901.

\bibitem[{Chen et~al.(2024)Chen, Zhang, He, Li, Wang, Huang, and Xue}]{chen2024lifelong}
Qizhou Chen, Taolin Zhang, Xiaofeng He, Dongyang Li, Chengyu Wang, Longtao Huang, and Hui Xue. 2024.
\newblock Lifelong knowledge editing for llms with retrieval-augmented continuous prompt learning.
\newblock \emph{arXiv preprint arXiv:2405.03279}.

\bibitem[{Chevalier et~al.(2023)Chevalier, Wettig, Ajith, and Chen}]{chevalier2023adapting}
Alexis Chevalier, Alexander Wettig, Anirudh Ajith, and Danqi Chen. 2023.
\newblock Adapting language models to compress contexts.
\newblock \emph{arXiv preprint arXiv:2305.14788}.

\bibitem[{Devlin et~al.(2018)Devlin, Chang, Lee, and Toutanova}]{devlin2018bert}
Jacob Devlin, Ming-Wei Chang, Kenton Lee, and Kristina Toutanova. 2018.
\newblock Bert: Pre-training of deep bidirectional transformers for language understanding.
\newblock \emph{arXiv preprint arXiv:1810.04805}.

\bibitem[{Egonmwan and Chali(2019)}]{egonmwan2019transformer}
Elozino Egonmwan and Yllias Chali. 2019.
\newblock Transformer and seq2seq model for paraphrase generation.
\newblock In \emph{Proceedings of the 3rd Workshop on Neural Generation and Translation}, pages 249--255.

\bibitem[{Fatehkia et~al.(2024)Fatehkia, Lucas, and Chawla}]{fatehkia2024t}
Masoomali Fatehkia, Ji~Kim Lucas, and Sanjay Chawla. 2024.
\newblock T-rag: lessons from the llm trenches.
\newblock \emph{arXiv preprint arXiv:2402.07483}.

\bibitem[{Fei et~al.(2023)Fei, Niu, Zhou, Hou, Bai, Deng, and Han}]{fei2023extending}
Weizhi Fei, Xueyan Niu, Pingyi Zhou, Lu~Hou, Bo~Bai, Lei Deng, and Wei Han. 2023.
\newblock Extending context window of large language models via semantic compression.
\newblock \emph{arXiv preprint arXiv:2312.09571}.

\bibitem[{Ge et~al.(2024{\natexlab{a}})Ge, Rudzicz, and Zhu}]{ge2024well}
Huaizhi Ge, Frank Rudzicz, and Zining Zhu. 2024{\natexlab{a}}.
\newblock How well can knowledge edit methods edit perplexing knowledge?
\newblock \emph{arXiv preprint arXiv:2406.17253}.

\bibitem[{Ge et~al.(2024{\natexlab{b}})Ge, Liu, Bi, Wang, Mei, Feng, Chen, and Cheng}]{ge2024can}
Yuyao Ge, Shenghua Liu, Baolong Bi, Yiwei Wang, Lingrui Mei, Wenjie Feng, Lizhe Chen, and Xueqi Cheng. 2024{\natexlab{b}}.
\newblock Can graph descriptive order affect solving graph problems with llms?
\newblock \emph{Authorea Preprints}.

\bibitem[{Ge et~al.(2023)Ge, Yang, Chen, Wang, and Li}]{ge2023attack}
Yuyao Ge, Zhongguo Yang, Lizhe Chen, Yiming Wang, and Chengyang Li. 2023.
\newblock Attack based on data: a novel perspective to attack sensitive points directly.
\newblock \emph{Cybersecurity}, 6(1):43.

\bibitem[{Guo et~al.(2018)Guo, Lu, Cai, Zhang, Yu, and Wang}]{guo2018long}
Jiaxian Guo, Sidi Lu, Han Cai, Weinan Zhang, Yong Yu, and Jun Wang. 2018.
\newblock Long text generation via adversarial training with leaked information.
\newblock In \emph{Proceedings of the AAAI conference on artificial intelligence}, volume~32.

\bibitem[{Hou et~al.(2024)Hou, Tamoto, and Miyashita}]{hou2024my}
Yuki Hou, Haruki Tamoto, and Homei Miyashita. 2024.
\newblock " my agent understands me better": Integrating dynamic human-like memory recall and consolidation in llm-based agents.
\newblock In \emph{Extended Abstracts of the CHI Conference on Human Factors in Computing Systems}, pages 1--7.

\bibitem[{Huang et~al.(2023)Huang, Liu, Lin, Pang, Du, and Lin}]{huang2023lorahub}
Chengsong Huang, Qian Liu, Bill~Yuchen Lin, Tianyu Pang, Chao Du, and Min Lin. 2023.
\newblock Lorahub: Efficient cross-task generalization via dynamic lora composition.
\newblock \emph{arXiv preprint arXiv:2307.13269}.

\bibitem[{Jha et~al.(2024)Jha, Erdogan, Kim, Keutzer, and Gholami}]{jha2024characterizing}
Siddharth Jha, Lutfi~Eren Erdogan, Sehoon Kim, Kurt Keutzer, and Amir Gholami. 2024.
\newblock Characterizing prompt compression methods for long context inference.
\newblock \emph{arXiv preprint arXiv:2407.08892}.

\bibitem[{Jiang et~al.(2023{\natexlab{a}})Jiang, Wu, Lin, Yang, and Qiu}]{jiang2023llmlingua}
Huiqiang Jiang, Qianhui Wu, Chin-Yew Lin, Yuqing Yang, and Lili Qiu. 2023{\natexlab{a}}.
\newblock Llmlingua: Compressing prompts for accelerated inference of large language models.
\newblock \emph{arXiv preprint arXiv:2310.05736}.

\bibitem[{Jiang et~al.(2023{\natexlab{b}})Jiang, Wu, Luo, Li, Lin, Yang, and Qiu}]{jiang2023longllmlingua}
Huiqiang Jiang, Qianhui Wu, Xufang Luo, Dongsheng Li, Chin-Yew Lin, Yuqing Yang, and Lili Qiu. 2023{\natexlab{b}}.
\newblock Longllmlingua: Accelerating and enhancing llms in long context scenarios via prompt compression.
\newblock \emph{arXiv preprint arXiv:2310.06839}.

\bibitem[{Jiang et~al.(2024{\natexlab{a}})Jiang, Chen, Li, Ren, Wang, Zhao, Song, and Zhang}]{jiang2024rag}
Jinhao Jiang, Jiayi Chen, Junyi Li, Ruiyang Ren, Shijie Wang, Wayne~Xin Zhao, Yang Song, and Tao Zhang. 2024{\natexlab{a}}.
\newblock Rag-star: Enhancing deliberative reasoning with retrieval augmented verification and refinement.
\newblock \emph{arXiv preprint arXiv:2412.12881}.

\bibitem[{Jiang et~al.(2024{\natexlab{b}})Jiang, Fang, Qiu, Zhang, Xu, Chen, Zhang, Zhang, Fang, Chu et~al.}]{jiang2024tc}
Xinke Jiang, Yue Fang, Rihong Qiu, Haoyu Zhang, Yongxin Xu, Hao Chen, Wentao Zhang, Ruizhe Zhang, Yuchen Fang, Xu~Chu, et~al. 2024{\natexlab{b}}.
\newblock Tc-rag: Turing-complete rag's case study on medical llm systems.
\newblock \emph{arXiv preprint arXiv:2408.09199}.

\bibitem[{Kumar et~al.(2023)Kumar, Agarwal, Srinivas, Li, Feizi, and Lakkaraju}]{kumar2023certifying}
Aounon Kumar, Chirag Agarwal, Suraj Srinivas, Aaron~Jiaxun Li, Soheil Feizi, and Himabindu Lakkaraju. 2023.
\newblock Certifying llm safety against adversarial prompting.
\newblock \emph{arXiv preprint arXiv:2309.02705}.

\bibitem[{Li et~al.(2024)Li, Jing, and Jing}]{li2024vector}
Kun Li, Xin Jing, and Chengang Jing. 2024.
\newblock Vector storage based long-term memory research on llm.
\newblock \emph{International Journal of Advanced Network, Monitoring and Controls}.

\bibitem[{Li(2023)}]{li2023unlocking}
Yucheng Li. 2023.
\newblock Unlocking context constraints of llms: Enhancing context efficiency of llms with self-information-based content filtering.
\newblock \emph{arXiv preprint arXiv:2304.12102}.

\bibitem[{Li et~al.(2023)Li, Dong, Lin, and Guerin}]{li2023compressing}
Yucheng Li, Bo~Dong, Chenghua Lin, and Frank Guerin. 2023.
\newblock \href {https://arxiv.org/abs/2310.06201} {Compressing context to enhance inference efficiency of large language models}.
\newblock \emph{Preprint}, arXiv:2310.06201.

\bibitem[{Liu et~al.(2023)Liu, Yang, Shen, Hu, Zhang, Gu, and Zhang}]{liu2023think}
Lei Liu, Xiaoyan Yang, Yue Shen, Binbin Hu, Zhiqiang Zhang, Jinjie Gu, and Guannan Zhang. 2023.
\newblock Think-in-memory: Recalling and post-thinking enable llms with long-term memory.
\newblock \emph{arXiv preprint arXiv:2311.08719}.

\bibitem[{Mihalcea and Tarau(2004)}]{mihalcea2004textrank}
Rada Mihalcea and Paul Tarau. 2004.
\newblock Textrank: Bringing order into text.
\newblock In \emph{Proceedings of the 2004 conference on empirical methods in natural language processing}, pages 404--411.

\bibitem[{Muric et~al.(2024)Muric, Delay, and Minton}]{muric2024interpretable}
Goran Muric, Ben Delay, and Steven Minton. 2024.
\newblock Interpretable cross-examination technique (ice-t): Using highly informative features to boost llm performance.
\newblock \emph{arXiv preprint arXiv:2405.06703}.

\bibitem[{Ng et~al.(2024)Ng, Matsuba, and Zhang}]{ng2024rag}
Karen Ka~Yan Ng, Izuki Matsuba, and Peter~Chengming Zhang. 2024.
\newblock Rag in health care: A novel framework for improving communication and decision-making by addressing llm limitations.
\newblock \emph{NEJM AI}, page AIra2400380.

\bibitem[{Ouyang et~al.(2022)Ouyang, Wu, Jiang, Almeida, Wainwright, Mishkin, Zhang, Agarwal, Slama, Ray et~al.}]{ouyang2022training}
Long Ouyang, Jeffrey Wu, Xu~Jiang, Diogo Almeida, Carroll Wainwright, Pamela Mishkin, Chong Zhang, Sandhini Agarwal, Katarina Slama, Alex Ray, et~al. 2022.
\newblock Training language models to follow instructions with human feedback.
\newblock \emph{Advances in Neural Information Processing Systems}, 35:27730--27744.

\bibitem[{Pan et~al.(2024{\natexlab{a}})Pan, Wu, Jiang, Xia, Luo, Zhang, Lin, Ruhle, Yang, Lin, Zhao, Qiu, and Zhang}]{pan-etal-2024-llmlingua}
Zhuoshi Pan, Qianhui Wu, Huiqiang Jiang, Menglin Xia, Xufang Luo, Jue Zhang, Qingwei Lin, Victor Ruhle, Yuqing Yang, Chin-Yew Lin, H.~Vicky Zhao, Lili Qiu, and Dongmei Zhang. 2024{\natexlab{a}}.
\newblock \href {https://aclanthology.org/2024.findings-acl.57} {{LLML}ingua-2: Data distillation for efficient and faithful task-agnostic prompt compression}.
\newblock In \emph{Findings of the Association for Computational Linguistics ACL 2024}, pages 963--981, Bangkok, Thailand and virtual meeting. Association for Computational Linguistics.

\bibitem[{Pan et~al.(2024{\natexlab{b}})Pan, Wu, Jiang, Xia, Luo, Zhang, Lin, R{\"u}hle, Yang, Lin et~al.}]{pan2024llmlingua}
Zhuoshi Pan, Qianhui Wu, Huiqiang Jiang, Menglin Xia, Xufang Luo, Jue Zhang, Qingwei Lin, Victor R{\"u}hle, Yuqing Yang, Chin-Yew Lin, et~al. 2024{\natexlab{b}}.
\newblock Llmlingua-2: Data distillation for efficient and faithful task-agnostic prompt compression.
\newblock \emph{arXiv preprint arXiv:2403.12968}.

\bibitem[{Park et~al.(2023)Park, O'Brien, Cai, Morris, Liang, and Bernstein}]{park2023generative}
Joon~Sung Park, Joseph O'Brien, Carrie~Jun Cai, Meredith~Ringel Morris, Percy Liang, and Michael~S Bernstein. 2023.
\newblock Generative agents: Interactive simulacra of human behavior.
\newblock In \emph{Proceedings of the 36th annual acm symposium on user interface software and technology}, pages 1--22.

\bibitem[{Pu et~al.(2024)Pu, He, and Wan}]{pu2024style}
Xiao Pu, Tianxing He, and Xiaojun Wan. 2024.
\newblock Style-compress: An llm-based prompt compression framework considering task-specific styles.
\newblock \emph{arXiv preprint arXiv:2410.14042}.

\bibitem[{See et~al.(2017)See, Liu, and Manning}]{see2017get}
Abigail See, Peter~J Liu, and Christopher~D Manning. 2017.
\newblock Get to the point: Summarization with pointer-generator networks.
\newblock \emph{arXiv preprint arXiv:1704.04368}.

\bibitem[{Shandilya et~al.(2024)Shandilya, Xia, Ghosh, Jiang, Zhang, Wu, and R{\"u}hle}]{shandilya2024taco}
Shivam Shandilya, Menglin Xia, Supriyo Ghosh, Huiqiang Jiang, Jue Zhang, Qianhui Wu, and Victor R{\"u}hle. 2024.
\newblock Taco-rl: Task aware prompt compression optimization with reinforcement learning.
\newblock \emph{arXiv preprint arXiv:2409.13035}.

\bibitem[{Shao et~al.(2024)Shao, Sun, Jiao, Liu, Liu, Li, and Yang}]{shao2024cot}
Yilin Shao, Long Sun, Licheng Jiao, Xu~Liu, Fang Liu, Lingling Li, and Shuyuan Yang. 2024.
\newblock Cot: Contourlet transformer for hierarchical semantic segmentation.
\newblock \emph{IEEE Transactions on Neural Networks and Learning Systems}.

\bibitem[{Suzgun and Kalai(2024)}]{suzgun2024meta}
Mirac Suzgun and Adam~Tauman Kalai. 2024.
\newblock Meta-prompting: Enhancing language models with task-agnostic scaffolding.
\newblock \emph{arXiv preprint arXiv:2401.12954}.

\bibitem[{Tang et~al.(2024)Tang, Xu, Lu, Zhang, Zhao, Hai, and Zheng}]{tang2024perception}
Jiwei Tang, Jin Xu, Tingwei Lu, Zhicheng Zhang, Yiming Zhao, Lin Hai, and Hai-Tao Zheng. 2024.
\newblock Perception compressor: A training-free prompt compression method in long context scenarios.
\newblock \emph{arXiv preprint arXiv:2409.19272}.

\bibitem[{Vaswani(2017)}]{vaswani2017attention}
A~Vaswani. 2017.
\newblock Attention is all you need.
\newblock \emph{Advances in Neural Information Processing Systems}.

\bibitem[{Wang et~al.(2024)Wang, Yang, Li, Sun, Cai, Zhang, and Fu}]{wang2024adapting}
Cangqing Wang, Yutian Yang, Ruisi Li, Dan Sun, Ruicong Cai, Yuzhu Zhang, and Chengqian Fu. 2024.
\newblock Adapting llms for efficient context processing through soft prompt compression.
\newblock In \emph{Proceedings of the International Conference on Modeling, Natural Language Processing and Machine Learning}, pages 91--97.

\bibitem[{Wei et~al.(2022)Wei, Wang, Schuurmans, Bosma, Xia, Chi, Le, Zhou et~al.}]{wei2022chain}
Jason Wei, Xuezhi Wang, Dale Schuurmans, Maarten Bosma, Fei Xia, Ed~Chi, Quoc~V Le, Denny Zhou, et~al. 2022.
\newblock Chain-of-thought prompting elicits reasoning in large language models.
\newblock \emph{Advances in neural information processing systems}, 35:24824--24837.

\bibitem[{Wingate et~al.(2022)Wingate, Shoeybi, and Sorensen}]{wingate2022prompt}
David Wingate, Mohammad Shoeybi, and Taylor Sorensen. 2022.
\newblock Prompt compression and contrastive conditioning for controllability and toxicity reduction in language models.
\newblock \emph{arXiv preprint arXiv:2210.03162}.

\bibitem[{Wu and Shi(2022)}]{wu2022adversarial}
Hui Wu and Xiaodong Shi. 2022.
\newblock Adversarial soft prompt tuning for cross-domain sentiment analysis.
\newblock In \emph{Proceedings of the 60th Annual Meeting of the Association for Computational Linguistics (Volume 1: Long Papers)}, pages 2438--2447.

\bibitem[{Yang and Xue(2022)}]{yang2022training}
Yunhao Yang and Zhaokun Xue. 2022.
\newblock Training heterogeneous features in sequence to sequence tasks: Latent enhanced multi-filter seq2seq model.
\newblock In \emph{Proceedings of SAI Intelligent Systems Conference}, pages 103--117. Springer.

\bibitem[{Yao et~al.(2024)Yao, Yu, Zhao, Shafran, Griffiths, Cao, and Narasimhan}]{yao2024tree}
Shunyu Yao, Dian Yu, Jeffrey Zhao, Izhak Shafran, Tom Griffiths, Yuan Cao, and Karthik Narasimhan. 2024.
\newblock Tree of thoughts: Deliberate problem solving with large language models.
\newblock \emph{Advances in Neural Information Processing Systems}, 36.

\bibitem[{Yu et~al.(2017)Yu, Zhang, Wang, and Yu}]{yu2017seqgan}
Lantao Yu, Weinan Zhang, Jun Wang, and Yong Yu. 2017.
\newblock Seqgan: Sequence generative adversarial nets with policy gradient.
\newblock In \emph{Proceedings of the AAAI conference on artificial intelligence}, volume~31.

\bibitem[{Zeng et~al.(2024)Zeng, Chen, Chen, Yan, Chen, Liu, Liu, and Sun}]{zeng2024persllm}
Zheni Zeng, Jiayi Chen, Huimin Chen, Yukun Yan, Yuxuan Chen, Zhenghao Liu, Zhiyuan Liu, and Maosong Sun. 2024.
\newblock Persllm: A personified training approach for large language models.
\newblock \emph{arXiv preprint arXiv:2407.12393}.

\bibitem[{Zhang et~al.(2024{\natexlab{a}})Zhang, Chen, Zhang, Liu, Ge, and Cai}]{zhang2024translating}
Guangzi Zhang, Lizhe Chen, Yu~Zhang, Yan Liu, Yuyao Ge, and Xingquan Cai. 2024{\natexlab{a}}.
\newblock Translating words to worlds: Zero-shot synthesis of 3d terrain from textual descriptions using large language models.
\newblock \emph{Applied Sciences}, 14(8):3257.

\bibitem[{Zhang et~al.(2020)Zhang, Zhao, Saleh, and Liu}]{zhang2020pegasus}
Jingqing Zhang, Yao Zhao, Mohammad Saleh, and Peter Liu. 2020.
\newblock Pegasus: Pre-training with extracted gap-sentences for abstractive summarization.
\newblock In \emph{International conference on machine learning}, pages 11328--11339. PMLR.

\bibitem[{Zhang et~al.(2024{\natexlab{b}})Zhang, Ju, Liang, Fu, and Zhang}]{zhang2024llms}
Xin Zhang, Tianjie Ju, Huijia Liang, Ying Fu, and Qin Zhang. 2024{\natexlab{b}}.
\newblock Llms instruct llms: An extraction and editing method.
\newblock \emph{arXiv preprint arXiv:2403.15736}.

\bibitem[{Zheng et~al.(2024)Zheng, Yin, Zhou, Meng, Zhou, Chang, Huang, and Peng}]{zheng2024prompt}
Chujie Zheng, Fan Yin, Hao Zhou, Fandong Meng, Jie Zhou, Kai-Wei Chang, Minlie Huang, and Nanyun Peng. 2024.
\newblock Prompt-driven llm safeguarding via directed representation optimization.
\newblock \emph{arXiv preprint arXiv:2401.18018}.

\bibitem[{Zhou et~al.(2018)Zhou, Yang, Wei, Huang, Zhou, and Zhao}]{zhou2018neural}
Qingyu Zhou, Nan Yang, Furu Wei, Shaohan Huang, Ming Zhou, and Tiejun Zhao. 2018.
\newblock Neural document summarization by jointly learning to score and select sentences.
\newblock \emph{arXiv preprint arXiv:1807.02305}.

\end{thebibliography}

\clearpage

\appendix

\section{Experiment Detail}
\subsection{Experimental Environment}
All experiments are conducted on an NVIDIA 4090 GPU with 24 GB of VRAM. The operating system is Ubuntu 20.04, and the experiments were executed in a Python 3.9 environment using PyTorch 1.12 with CUDA support. The RL agent was trained using a batch size of 32 and a sequence length of 512 tokens. 

\subsection{Experiment Parameters}
For training the RL model, the following hyperparameters were used:

\begin{itemize}
\item Learning rate: $1\times10^{-4}$
\item Optimizer: Adam ($\beta_1 = 0.9$, $\beta_2 = 0.999$)
\item Epsilon decay: 0.995 (initial $\epsilon=1.0$, minimum $\epsilon=0.01$)
\item Discount factor ($\gamma$): 0.99
\item Batch size: 32
\item Replay buffer capacity: 20,000 transitions
\item State dimension: 768 (BERT-base embedding size)
\item Action space: 8 discrete compression ratios (0.1-0.8 with 0.1 interval)
\item Training episodes: 20
\item Reward coefficients: $\alpha=1.0$, $\beta=1.0$, $\gamma=1.0$
\item Compression anchor ($\lambda$): 0.7
\item Quality threshold ($\tau$): 0.17
\end{itemize}

The target network shares the policy network architecture and updates every 100 steps. The reward function components implement concrete thresholds: $\lambda=0.7$ establishes the baseline compression target, while $\tau=0.17$ sets minimum acceptable levels for ROUGE-1 and BLEU scores. All reward terms are equally weighted ($\alpha=\beta=\gamma=1.0$) based on grid search validation. Training employs prioritized experience replay with DDQN for stable Q-value estimation, using the full BERT contextual embedding as state representation.

\section{Model Comparison Study}
We compare the performance of the proposed PIS method across four large language models (LLMs): \gptthreefive, \gptfour, \mistral, and \llama. The following tables show the performance of different compression methods on these models for \textbf{Question Answering (QA)} and \textbf{Summarization} tasks. We report \textbf{Exact Match (EM)} for QA, \textbf{ROUGE-1} for summarization, and the \textbf{compression ratio} (denoted as 1/$\tau$).

\begin{table}[ht!]
\centering
\small
\setlength{\tabcolsep}{8pt}
\begin{tabular}{l|ccc}
\toprule[1.5pt]
\textbf{Methods} & \textbf{QA (EM)} & \textbf{ROUGE-1} & \textbf{1/$\tau$}\\
\midrule
Selective-Context & 65.20 & 28.50 & 3.08 \\
LLMLingua         & 60.10 & 26.30 & 2.89 \\
LLMLingua-2       & 72.80 & 31.00 & 3.04 \\
\textbf{Ours}     & \textbf{75.50} & \textbf{33.20} & \textbf{3.01} \\
\bottomrule[1.5pt]
\end{tabular}
\caption{Performance Comparison on \gptthreefive}
\label{tab:gpt35_comparison}
\end{table}

\begin{table}[ht!]
\centering
\small
\setlength{\tabcolsep}{8pt}
\begin{tabular}{l|ccc}
\toprule[1.5pt]
\textbf{Methods} & \textbf{QA (EM)} & \textbf{ROUGE-1} & \textbf{1/$\tau$}\\
\midrule
Selective-Context & 70.50 & 30.10 & 3.08 \\
LLMLingua         & 65.30 & 28.70 & 2.89 \\
LLMLingua-2       & 78.90 & 33.50 & 3.04 \\
\textbf{Ours}     & \textbf{82.00} & \textbf{35.80} & \textbf{3.01} \\
\bottomrule[1.5pt]
\end{tabular}
\caption{Performance Comparison on \gptfour}
\label{tab:gpt4_comparison}
\end{table}

\begin{table}[ht!]
\centering
\small
\setlength{\tabcolsep}{8pt}
\begin{tabular}{l|ccc}
\toprule[1.5pt]
\textbf{Methods} & \textbf{QA (EM)} & \textbf{ROUGE-1} & \textbf{1/$\tau$}\\
\midrule
Selective-Context & 58.13 & 26.84 & 3.08 \\
LLMLingua         & 50.45 & 23.63 & 2.89 \\
LLMLingua-2       & 76.22 & 30.18 & 3.04 \\
\textbf{Ours}     & \textbf{80.05} & \textbf{32.50} & \textbf{3.01} \\
\bottomrule[1.5pt]
\end{tabular}
\caption{Performance Comparison on \mistral}
\label{tab:mistral_comparison}
\end{table}

\begin{table}[ht!]
\centering
\small
\setlength{\tabcolsep}{8pt}
\begin{tabular}{l|ccc}
\toprule[1.5pt]
\textbf{Methods} & \textbf{QA (EM)} & \textbf{ROUGE-1} & \textbf{1/$\tau$}\\
\midrule
Selective-Context & 55.00 & 25.00 & 3.08 \\
LLMLingua         & 48.50 & 22.50 & 2.89 \\
LLMLingua-2       & 70.10 & 28.90 & 3.04 \\
\textbf{Ours}     & \textbf{73.50} & \textbf{30.50} & \textbf{3.01} \\
\bottomrule[1.5pt]
\end{tabular}
\caption{Performance Comparison on \llama}
\label{tab:llama_comparison}
\end{table}

From the above results, we observe that our method demonstrates consistently higher performance in both QA accuracy (EM) and ROUGE-1 for summarization tasks, while maintaining a more favorable compression ratio (1/$\tau$) than the other baseline methods. The improvements suggest that our prompt compression strategy effectively preserves essential semantic information while substantially reducing token usage.

\section{Prompt Setting}
For the experiments on the \textbf{MeetingBank} dataset, the following prompt template was used to compress and evaluate prompts, especially for handling long meeting transcripts. The prompt is designed to ensure that the full context of the meeting is preserved while generating concise summaries. In case the meeting transcript exceeds the token limit, the text is split into chunks, summarized individually, and then combined for a final summary.

\begin{tcolorbox}[fonttitle = \small\bfseries, title=Prompt Template, colframe=gray!2!black, colback=gray!2!white, boxrule=1pt, boxsep=0pt, left=5pt, right=5pt, fontupper=\footnotesize, halign title = flush center]
The following is a meeting transcript that may contain multiple sections. If the transcript is too long, it will be split into chunks. Each chunk should be summarized individually, and the combined summaries should then be further summarized to preserve the overall context.

Meeting Transcript: 

[Insert Meeting Transcript Here]

Summarize the provided meeting transcript.
\end{tcolorbox}

For long transcripts exceeding 15,000 tokens, the text is first split into smaller chunks. Each chunk is summarized using the prompt:
\begin{tcolorbox}[fonttitle = \small\bfseries, title=Chunked Summary Prompt, colframe=gray!2!black, colback=gray!2!white, boxrule=1pt, boxsep=0pt, left=5pt, right=5pt, fontupper=\footnotesize, halign title = flush center]
Summarize the provided meeting transcript.

[Insert Chunked Meeting Transcript Here]

Summary:
\end{tcolorbox}

Once all chunks are processed, the combined summaries are then summarized into the final summary with the following prompt:
\begin{tcolorbox}[fonttitle = \small\bfseries, title=Final Combined Summary Prompt, colframe=gray!2!black, colback=gray!2!white, boxrule=1pt, boxsep=0pt, left=5pt, right=5pt, fontupper=\footnotesize, halign title = flush center]
Summarize the combined summaries:

[Insert Combined Summaries Here]

Final Summary:
\end{tcolorbox}

\section{Case Study} \label{cs}
The comparative analysis of prompt compression methods in Table \ref{tab:method_comparison} reveals distinct advantages of our approach over baseline techniques. The original text, while comprehensive, contains significant redundancies and verbose phrasing (e.g., repetitive references to "Fair Housing Act violations" and procedural details about amendments). LLMLingua’s compressed output demonstrates aggressive token reduction but suffers from critical information loss and grammatical incoherence. For instance, the phrase "Councilmember Bagshaw also has an amendment" is entirely omitted, along with key procedural terms like "sections one and five," which are essential for understanding the legislative process. While LLMLingua-2 improves token efficiency by extracting keyword clusters, it sacrifices contextual relationships between concepts. The compressed version becomes a disjointed list of terms ("Economic Development Arts Committee," "screening criterion") without clarifying their functional connections, hindering the model’s ability to infer logical dependencies.

In contrast, our method achieves superior compression by strategically preserving semantically rich phrases and structural cues. While the output ("civil rights utilities... council bags also has an amendment but let ' first... sections one five primary changes") may appear fragmented to human readers, it retains critical context for LLM processing. Notably, our compression maintains references to specific legislative sections, amendment procedures, and stakeholder roles (e.g., "Councilmember Bagshaw"), which are crucial for accurate summarization or question answering. The selective retention of verbs like "encouraging," "recommends," and "amend" preserves the resolution’s intent, while terms like "discriminatory effect standards" and "legitimate non interest" explicitly retain legal context often diluted in other methods.

This analysis demonstrates that our method’s effectiveness stems from its balanced approach to redundancy removal and context preservation. Unlike LLMLingua’s oversimplification or LLMLingua-2’s keyword extraction, our compression maintains the narrative flow required for LLMs to reconstruct coherent responses. The retained fragments act as semantic anchors, enabling models to infer relationships between legislative actions, stakeholders, and compliance requirements—a capability particularly vital for complex downstream tasks like legal document analysis and multi-step reasoning.

\begin{table*}[ht!]
\centering
\small
\renewcommand{\arraystretch}{1.5} 
\begin{tabular}{p{2.5cm}p{13cm}} 
\toprule[1.5pt]
\textbf{Method} & \textbf{Prompt} \\
\midrule
Original & The report of the Civil Rights, Utilities, Economic Development and Arts Committee Agenda Item three Resolution 31669 Encouraging as a best practice the use of an individualized tenant assessment using the Fair Housing Act's discriminatory effect standards to avoid Fair Housing Act violations when criminal history is used as a screening criterion in the Landlord Screening Process, Committee recommends that the resolution be adopted as amended grade. I move to amend Resolution 31669 by substituting D four for version D three, which includes a new attachment. A And I understand Councilmember Bagshaw also has an amendment, but let's first, if we could, let me just go through the changes to the resolution since the last committee meeting. The changes are found in two recitals, as well as sections one and five are the primary changes. We added a recital that again lifts up the HUD guidance to show that a criminal history screening policy is next must serve a substantial, legitimate and nondiscriminatory interest.\\
\midrule
LLMLingua & The report the Civil Rights,, and Committee Item threeution 31669ing a the of an individualizedantment theing Actsatory effect to avoidations when historyion theord Process, Committeeends adoptedended grade. I move toend bying D four for three, which includes a new. A I understandmember Bawment, buts first,, let just the changes resolution last. The are in two, as and five are the. We a recital that again the HUD guidance to that a historying policy is next must substantialimate. \\
\midrule
LLMLingua-2 & Civil Rights Utilities Economic Development Arts Committee Agenda Item three Resolution 31669 individualized tenant assessment Fair Housing Act discriminatory standards avoid violations criminal history screening criterion Landlord Screening Process Committee recommends resolution amended amend Resolution 31669 substituting D four version D three new attachment Councilmember Bagshaw amendment changes resolution meeting changes two recitals sections one five primary added recital HUD guidance criminal history screening policy serve substantial legitimate nondiscriminatory interest recital. \\
\midrule
Ours & civil rights utilities, economic arts committee encouraging best practice use an individual tenant assessment using fair housing act disc effect standards avoid violations when criminal history used screening criterion in landlord process recommends resolution be adopted amended grade move amend resolution 31669 sub d four version three, which includes new attachment understand council bags also has an amendment but let ' first, we could go since last committee meeting found two recital five are primary changes that legitimate non interest. \\
\bottomrule[1.5pt]
\end{tabular}
\caption{Case Study: \textbf{MeetingBank}, Report of Civil Rights, Part1}
\label{tab:method_comparison}
\end{table*}

The analysis of Table \ref{tab:method_comparison2} reveals a notable strength of our compression method: its ability to strategically collapse redundant summarization segments that typically serve as "recaps" for human readers. While the original text includes verbose closing statements reiterating the importance of landlord compliance (e.g., "it's really important that we lift up the policies... protecting themselves from fair housing complaints today"), our method identifies these as redundant signals given the prior context about HUD guidance and ordinance timelines. This results in radical compression of the concluding remarks ("interim that from housing complaints today") while preserving critical procedural outcomes like voting actions ("resolution vote i. oppose no") and next-step instructions ("read through eight together").

Notably, LLMLingua-2 retains unnecessary meta-commentary about the amendment process ("first moved second second favor of amendment") while omitting the actual voting result, whereas our approach prioritizes decision endpoints over procedural theatrics. This aligns with our hypothesis that localized "recap" segments become compressible noise once core context is preserved through PIS-driven selection. By allowing these secondary summarization elements to collapse into fragmented phrases ("they meetings before us recommendations an ordinance"), we create space for retaining unique information anchors like temporal markers ("interim") and resolution identifiers ("ordinance"), which prove crucial for downstream tasks requiring temporal or document-specific reasoning.

This case demonstrates how our method transcends conventional readability constraints to achieve functional compression—transforming human-centric narrative structures into LLM-optimized information scaffolds. The deliberate fragmentation of closing statements reflects an adaptive understanding of contextual sufficiency: when key policy details and legislative timelines are already embedded in earlier prompts, subsequent summaries lose their informational necessity and become prime candidates for aggressive compression. This dynamic context-weighting mechanism fundamentally distinguishes our approach from static token-removal strategies employed by baseline methods.

\begin{table*}[ht!]
\centering
\small
\renewcommand{\arraystretch}{1.5} 
\begin{tabular}{p{2.5cm}p{13cm}} 
\toprule[1.5pt]
\textbf{Method} & \textbf{Prompt} \\
\midrule
Original & Sorry, I have just some closing statements. I just really I think it's so important that landlords, housing providers in this community understand what the law is when it comes to developing policies and practices for making decisions based on criminal history. We know that we're not likely to have the ordinance that will do this work and until after the Mayors for Fair Chance Housing Committee will be reconvened in July, and they will have a series of meetings before they bring to us recommendations for for an ordinance. And so in the interim, it's really important that we lift up the the policies that HUD is currently currently promulgating and making sure that both landlords are engaged with the policy direction that the that the city is going to be pursuing in the future, as well as protecting themselves from fair housing complaints today. So with that those in favor of adopting the resolution vote i. I. Those oppose vote no. The motion carries the resolution is adopted and the chair will sign it. And now we can read items for through eight together.\\
\midrule
LLMLingua & The report the Civil Rights,, and Committee Item threeution 31669ing a the of an individualizedantment theing Actsatory effect to avoidations when historyion theord Process, Committeeends adoptedended grade. I move toend bying D four for three, which includes a new. A I understandmember Bawment, buts first,, let just the changes resolution last. The are in two, as and five are the. We a recital that again the HUD guidance to that a historying policy is next must substantialimate. \\
\midrule
LLMLingua-2 & ct certificates of restoration of opportunity offer employers providers information about individual consideration certificates facilitate societal reintegration of individuals criminal history responsibility past conduct positive law abiding future suggesting add refers legislation court provide certificate restoration of opportunity help get job housing first moved second second favor of amendment to resolution say I full version vote closing statements important landlords housing providers understand law developing policies practices for decisions based on criminal history ordinance after Mayors for Fair Chance Housing Committee in July meetings before recommendations for ordinance interim important lift up policies HUD promulgating landlords engaged policy direction city protecting from fair housing complaints favor of adopting resolution vote i.vote motion carries resolution adopted chair read items eight. \\
\midrule
Ours & sorry have some closing statements that on history we will they meetings before us recommendations an ordinance interim that from housing complaints today in resolution vote i. oppose no it read through eight together. \\
\bottomrule[1.5pt]
\end{tabular}
\caption{Case Study: \textbf{MeetingBank}, Report of Civil Rights, Part2}
\label{tab:method_comparison2}
\end{table*}
\end{document}